\documentclass{article}
\usepackage{ijcai19}

\usepackage{times}
\usepackage{soul}
\usepackage{url}
\usepackage[hidelinks]{hyperref}
\usepackage[utf8]{inputenc}
\usepackage[small]{caption}
\usepackage{graphicx}
\usepackage{amsmath}
\usepackage{amssymb}
\usepackage{booktabs}
\usepackage{epsfig}
\usepackage{microtype}      % microtypograph
\usepackage[american]{babel}

\newcommand{\eg}{\emph{e.g.}}
\newcommand{\ie}{\emph{i.e.}}

\graphicspath{{figure/}}

\title{Deep learning for fine-grained image analysis: A survey}

\author{
Xiu-Shen Wei$^1$\and
Jianxin Wu$^2$\and
Quan Cui$^{1,3}$\\
\affiliations
$^1$Megvii Research Nanjing, Megvii Technology, Nanjing, China\\
$^2$National Key Laboratory for Novel Software Technology, Nanjing University, Nanjing, China\\
$^3$Graduate School of IPS, Waseda University, Fukuoka, Japan\\
}

\begin{document}

\maketitle

\begin{abstract}
Computer vision (CV) is the process of using machines to understand and analyze imagery, which is an integral branch of artificial intelligence. Among various research areas of CV, fine-grained image analysis (FGIA) is a longstanding and fundamental problem, and has become ubiquitous in diverse real-world applications. The task of FGIA targets analyzing visual objects from subordinate categories, \eg, species of birds or models of cars. The small inter-class variations and the large intra-class variations caused by the fine-grained nature makes it a challenging problem. During the booming of deep learning, recent years have witnessed remarkable progress of FGIA using deep learning techniques. In this paper, we aim to give a survey on recent advances of deep learning based FGIA techniques in a systematic way. Specifically, we organize the existing studies of FGIA techniques into three major categories: fine-grained image recognition, fine-grained image retrieval and fine-grained image generation. In addition, we also cover some other important issues of FGIA, such as publicly available benchmark datasets and its related domain specific applications. Finally, we conclude this survey by highlighting several directions and open problems which need be further explored by the community in the future.
\end{abstract}

\section{Introduction}

Computer vision (CV) is an interdisciplinary scientific field of artificial intelligence (AI), which deals with how computers can be made to gain high-level understanding from digital images or videos. The tasks of computer vision include methods for acquiring, processing, analyzing and understanding digital images, and the process of extracting numerical or symbolic information, \eg, in the forms of decisions or predictions, from high-dimensional raw image data in the real world.

\begin{figure}[t!]
\centering
{\includegraphics[width=\columnwidth]{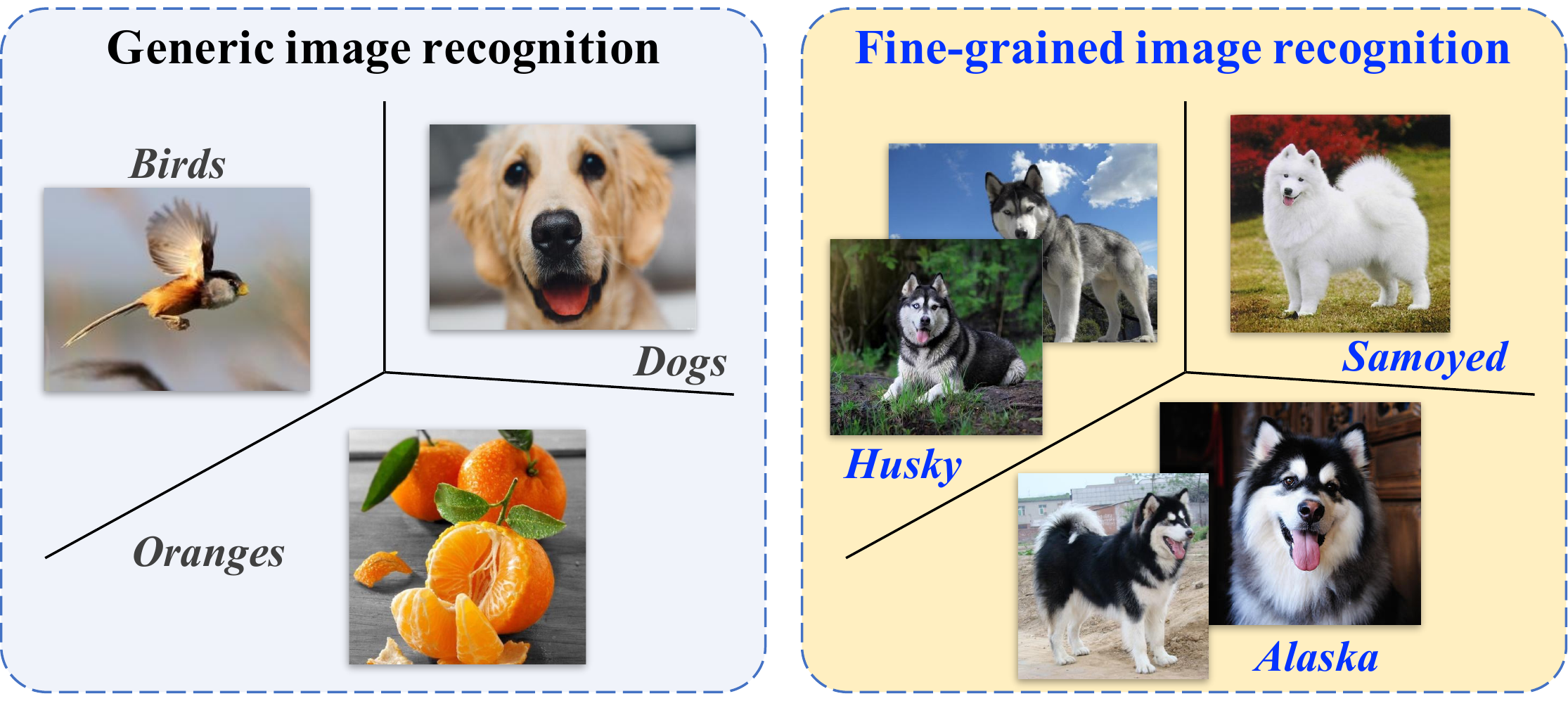}}
%\vspace{-0.5em}
\caption{Fine-grained image analysis \emph{vs}. generic image analysis (taking the recognitiont task for an example).}
\label{fig:fgvsgeneric}
\end{figure}

As an interesting, fundamental and challenging problem in computer vision, fine-grained image analysis (FGIA) has been an active area of research for several decades. The goal of FGIA is to retrieve, recognize and generate images belonging to multiple subordinate categories of a super-category (\emph{aka} meta-category), \eg, different species of animals/plants, different models of cars, different kinds of retail products, etc (cf. Fig.~\ref{fig:fgvsgeneric}). In the real-world, FGIA enjoys a wide-range of applications in both industry and research societies, such as automatic biodiversity monitoring, climate change evaluation, intelligent retail, intelligent transportation, and many more. Particularly, a number of influential academic competitions about FGIA are frequently held on \texttt{Kaggle}.\footnote{Kaggle is an online community of data scientists and machine learners: \url{https://www.kaggle.com/}.} Several representative competitions, to name a few, are the Nature Conservancy Fisheries Monitoring (for fish species categorization), Humpback Whale Identification (for whale identity categorization) and so on. Each competition attracted more than 300 teams worldwide to participate, and some even exceeded 2,000 teams.

On the other hand, deep learning techniques~\cite{natureDL} have emerged in recent years as powerful methods for learning feature representations directly from data, and have led to remarkable breakthroughs in the filed of FGIA. With rough statistics on each year, on average, there are around ten conference papers of deep learning based FGIA techniques published on each of AI's and CV's premium conferences, like IJCAI, AAAI, CVPR, ICCV, ECCV, etc. It shows that FGIA with deep learning is of notable research interests. Given this period of rapid evolution, the aim of this paper is to provide a comprehensive survey of the recent achievements in the FGIA filed brought by deep learning techniques.

In the literature, there was an existing survey related to fine-grained tasks, \ie, \cite{bozhao_ijac_2017}, which simply included several fine-grained \emph{recognition} approaches for comparisons. Our work differs with it in that ours is more comprehensive. Specifically, except for fine-grained recognition, we also analyze and discuss the other two central fine-grained {analysis} tasks, \ie, fine-grained image \emph{retrieval} and fine-grained image \emph{generation}, which can not be overlooked as they are two integral aspects of FGIA. Additionally, on another important AI conference in the Pacific Rim nations, PRICAI, Wei and Wu organized a specific tutorial\footnote{\url{http://www.weixiushen.com/tutorial/PRICAI18/FGIA.html}} aiming at the fine-grained image analysis topic. We refer interested readers to the tutorial which provides some additional detailed information.

\begin{figure}[t!]
\centering
{\includegraphics[width=\columnwidth]{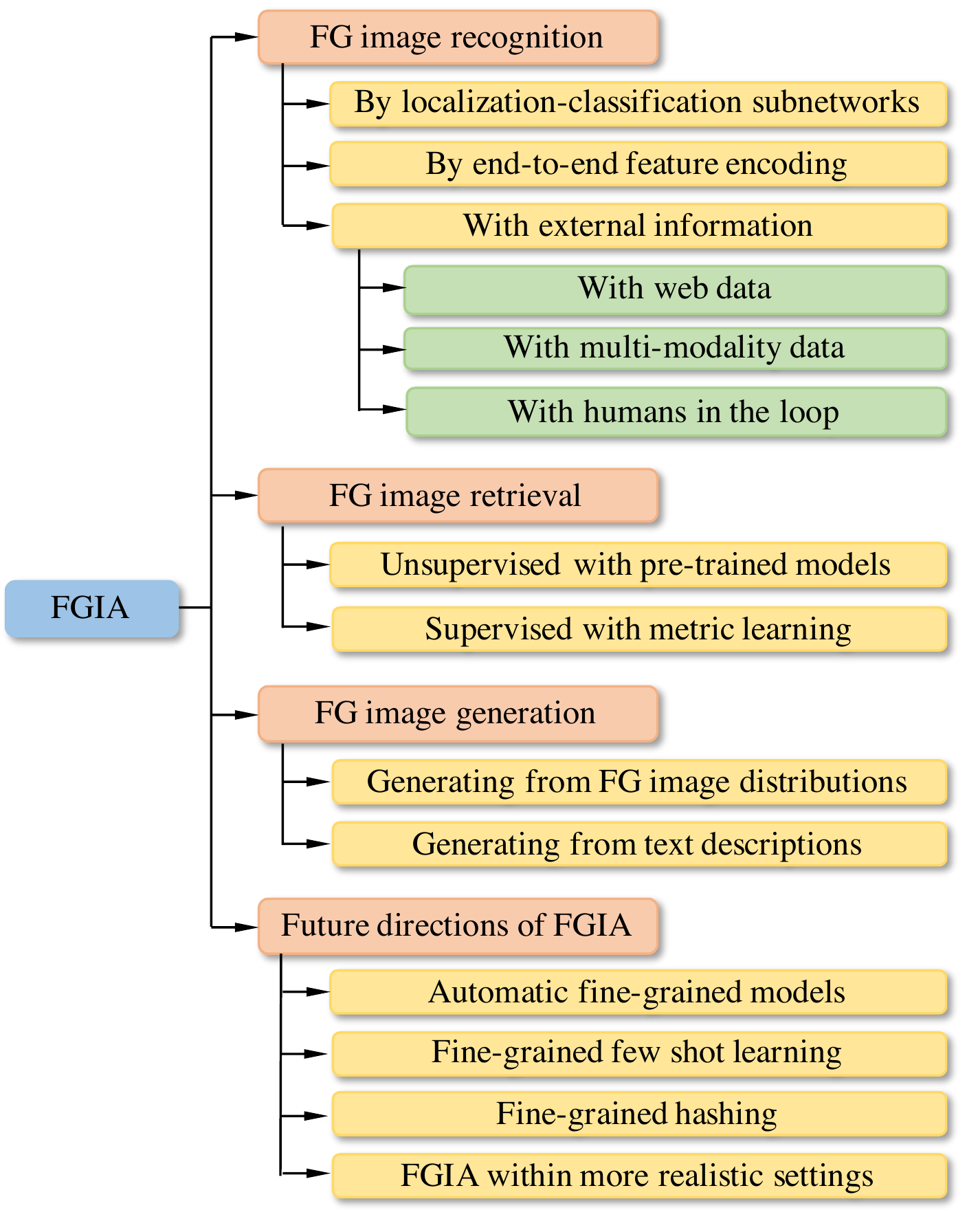}}
%\vspace{-0.5em}
\caption{Main aspects of our hierarchical and structrual organization of fine-grained image analysis (FGIA) in this survey paper.}
\label{fig:structure}
\end{figure}

In this paper, our survey take a unique deep learning based perspective to review the recent advances of FGIA in a systematic and comprehensive manner. The main contributions of this survey are three-fold:
\begin{itemize}
\item We give a comprehensive review of FGIA techniques based on deep learning, including problem backgrounds, benchmark datasets, a family of FGIA methods with deep learning, domain-specific FGIA applications, etc.
\item We provide a systematic overview of recent advances of deep learning based FGIA techniques in a hierarchical and structural manner, cf. Fig.~\ref{fig:structure}.
\item We discuss the challenges and open issues, and identify the new trends and future directions to provide a potential road map for fine-grained researchers or other interested readers in the broad AI community.
\end{itemize}

The rest of the survey is organized as follows. Section~\ref{sec:background} introduce backgrounds of this paper, \ie, the FGIA problem and its main challenges. In Section~\ref{sec:datasets}, we review multiple commonly used fine-grained benchmark datasets. Section~\ref{sec:fgrecognition} analyzes the three main paradigms of fine-grained image recognition. Section~\ref{sec:fgretrieval} presents recent progress of fine-grained image retrieval. Section~\ref{sec:fggeneration} discusses fine-grained image generation from a generative perspective. Furthermore, in Section~\ref{sec:otherapp}, we introduce some other domain specific applications of real-world related to FGIA. Finally, we conclude this paper and discuss future directions and open issues in Section~\ref{sec:conclusion}.

\section{Background: problem and main challenges}\label{sec:background}

\begin{figure}[t!]
\centering
{\includegraphics[width=0.95\columnwidth]{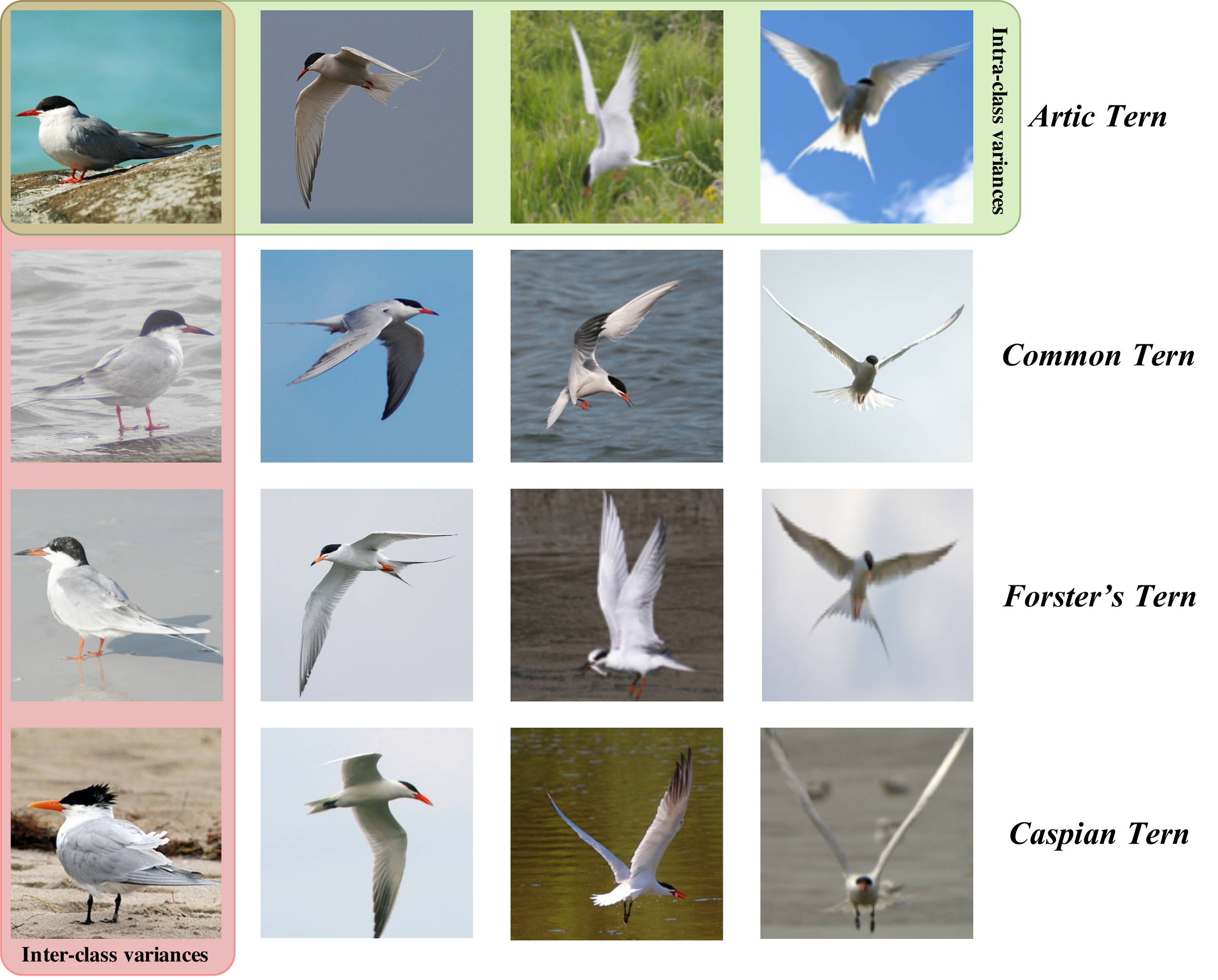}}
%\vspace{-1em}
\caption{Key challenge of fine-grained image analysis, \ie, small inter-class variations and large intra-class variations. We here present each of four \texttt{Tern} species in each row in the figure, respectively.}
\label{fig:cub_example}
\end{figure}

In this section, we summarize the related background of this paper, including the problem and its key challenges.

%\subsection{The problem}

Fine-grained image analysis (FGIA) focuses on dealing with the objects belonging to multiple \emph{sub-categories} of the same meta-category (\eg, birds, dogs and cars), and generally involves central tasks like fine-grained image recognition, fine-grained image retrieval, fine-grained image generation, etc.

What distinguishes FGIA from the generic one is: in generic image analysis, the target objects belong to coarse-grained meta-categories (\eg, birds, oranges and dogs), and thus are visually quite different. However, in FGIA, since objects come from sub-categories of one meta-category, the fine-grained nature causes them visually quite similar. We take image recognition for illustration. As shown in Fig.~\ref{fig:fgvsgeneric}, in fine-grained recognition, the task is required to identify multiple similar species of dogs, \eg, \texttt{Husky}, \texttt{Samoyed} and \texttt{Alaska}. For accurate recognition, it is desirable to distinguish them by capturing slight and subtle differences (\eg, ears, noses, tails), which also meets the demand of other FGIA tasks (\eg, retrieval and generation).

\begin{figure*}[t!]
\centering
{\includegraphics[width=0.98\textwidth]{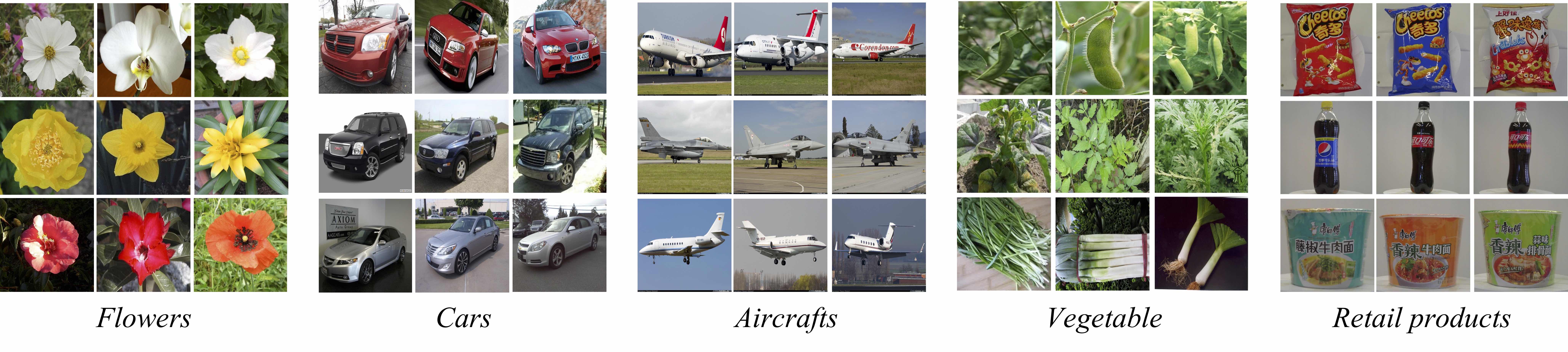}}
\vspace{-1em}
\caption{Example fine-grained images belonging to different species of flowers/vegetable, different models of cars/aircrafts and different kinds of retail products. Accurate identification of these fine-grained objects requires the dependences on the discriminative but subtle object parts or image regions. (Best viewed in color and zoomed in.)}
\label{fig:datasetdemo}
\end{figure*}

\begin{table*}[h]
\centering
\small 
\caption{Summary of popular fine-grained image datasets. Note that ``BBox'' indicates whether this dataset provides object bounding box supervisions. ``Part anno.'' means providing the key part localizations. ``HRCHY'' corresponds to hierarchical labels. ``ATR'' represents the attribute labels (\eg, wing color, male, female, etc). ``Texts'' indicates whether fine-grained text descriptions of images are supplied.}
%\vspace{-1em}
	\setlength{\tabcolsep}{3.6pt}
	\begin{tabular}{|c||c|c|c|c|c|c|c|c|}
		\hline
		Dataset name & Meta-class & $\sharp$ images & $\sharp$ categories  &BBox & Part anno. & HRCHY& ATR & Texts \\
		\hline \hline 
		\textit{Oxford Flower} \cite{Flowers08} 	 & Flowers & ~~~~8,189 & ~~102  &  &  & &  & \checkmark \\
		\textit{CUB200-2011}  \cite{WahCUB200_2011}  & Birds   & ~~11,788 & ~~200  & \checkmark & \checkmark &  & \checkmark & \checkmark\\
		\textit{Stanford Dog} \cite{Khosla11stanforddogs} & Dogs & ~~20,580 & ~~120  & \checkmark & &  &   & \\
		\textit{Stanford Car}  \cite{cars} & Cars & ~~16,185 & ~~196  & \checkmark &  & &  & \\
		\textit{FGVC Aircraft} \cite{airplanes}	& Aircrafts  & ~~10,000 & ~~100   & \checkmark & &\checkmark &  & \\
		\textit{Birdsnap} \cite{Birdsnap14} & Birds & ~~49,829 & ~~500  & \checkmark & \checkmark & & \checkmark &  \\
		\textit{Fru92} \cite{vegfru} & Fruits & ~~69,614 & ~~~~92  &  &  & \checkmark&  &  \\
		\textit{Veg200} \cite{vegfru} & Vegetable & ~~91,117 & ~~200  &  &  & \checkmark &  &  \\
		\textit{iNat2017} \cite{inat2017} & Plants \& Animals & 859,000 & 5,089  & \checkmark &  & \checkmark &  &  \\
		\textit{RPC} \cite{rpc} & Retail products & ~~83,739 & ~~200  & \checkmark &  & \checkmark &  &  \\
		\hline
	\end{tabular}
	\label{table:fgdataset} 
\end{table*}

Furthermore, fine-grained nature also brings the \emph{small inter-class variations} caused by highly similar sub-categories, and the \emph{large intra-class variations} in poses, scales and rotations, as presented by Fig.~\ref{fig:cub_example}. It is the opposite of the generic image analysis (\ie, the small intra-class variations and the large inter-class variations), which makes fine-grained image analysis a challenging problem.

\section{Benchmark datasets}\label{sec:datasets}

\begin{figure}[t!]
\centering
{\includegraphics[width=\columnwidth]{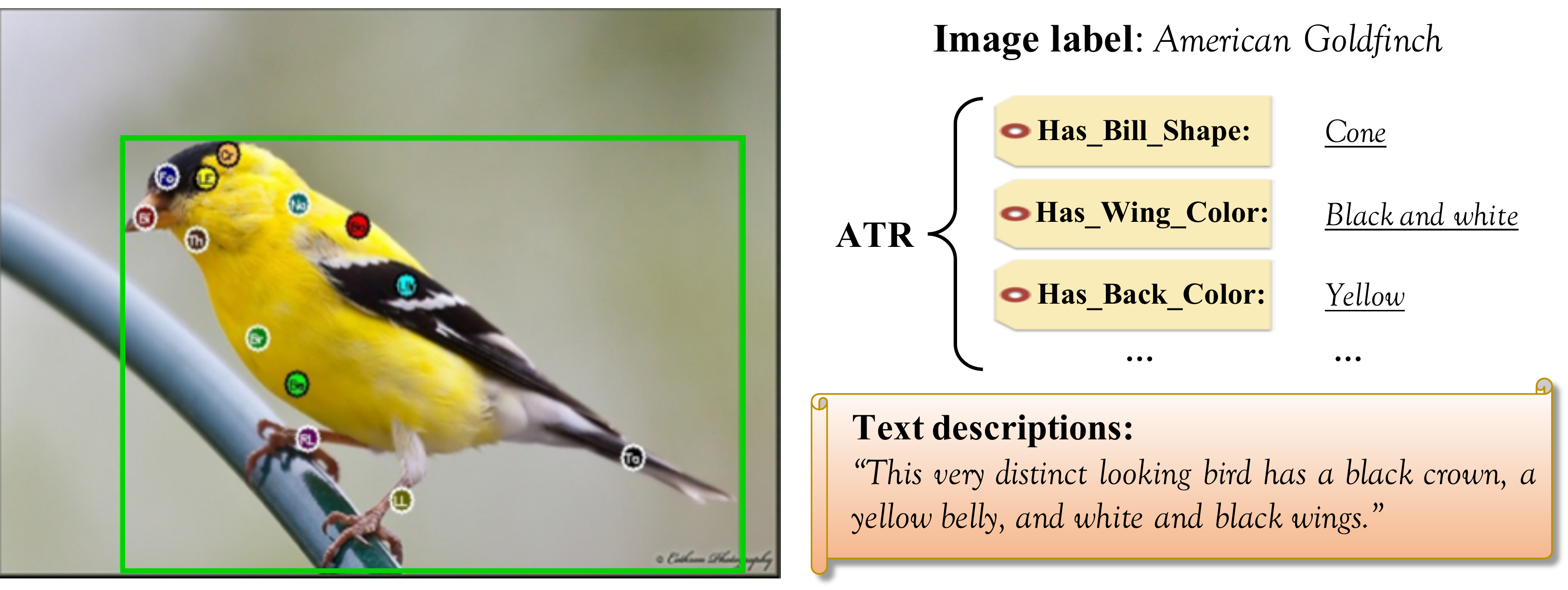}}
%\vspace{-1.6em}
\caption{An example image with its supervisions associated with \emph{CUB200-2011}. As shown, multiple types of supervisions include: image labels, part annotations (\emph{aka} key point localizations), object bounding boxes (\ie, the green one), attribute labels (\ie, ``ATR''), and text descriptions by natural languages. (Best viewed in color.)}
\label{fig:supervisions}
\end{figure}

In the past decade, the vision community has released many benchmark fine-grained datasets covering diverse domains such as birds~\cite{WahCUB200_2011,Birdsnap14}, dogs~\cite{Khosla11stanforddogs}, cars~\cite{cars}, airplanes~\cite{airplanes}, flowers~\cite{Flowers08}, vegetable~\cite{vegfru}, fruits~\cite{vegfru}, retail products~\cite{rpc}, etc (cf. Fig.~\ref{fig:datasetdemo}). In Table~\ref{table:fgdataset}, we list a number of image datasets commonly used by the fine-grained community, and specifically indicate their meta-category, the amounts of fine-grained images, the number of fine-grained categories, extra different kinds of available supervisions, \ie, bounding boxes, part annotations, hierarchical labels, attribute labels and text visual descriptions, cf. Fig.~\ref{fig:supervisions}.

These datasets have been one of the most important factors for the considerable progress in the filed, not only as a common ground for measuring and comparing performance of competing approaches, but also pushing this filed towards increasingly complex, practical and challenging problems.

Specifically, among them, \emph{CUB200-2011} is one of the most popular fine-grained datasets. Almost all the FGIA approaches choose it for comparisons with state-of-the-arts. Moreover, constant contributions are made upon \emph{CUB200-2011} for further research, \eg, collecting text descriptions of the fine-grained images for multi-modality analysis, cf.~\cite{fgtextcvpr2016,yuxinpengcvpr2017}.

Additionally, in recent years, more challenging and practical fine-grained datasets are proposed increasingly, \eg, \emph{iNat2017} for natural species of plants, animals~\cite{inat2017} and \emph{RPC} for daily retail products~\cite{rpc}. Many novel features deriving from these datasets are, to name a few, large-scale, hierarchical structure, domain gap and long-tail distribution, which reveals the practical requirements in real-world and could arouse the studies of FGIA in more realistic settings.

\section{Fine-grained image recognition}\label{sec:fgrecognition}

Fine-grained image recognition has been the most active research area of FGIA in the past decade. In this section, we review the milestones of fine-grained recognition frameworks since deep learning entered the filed. Broadly, these fine-grained recognition approaches can be organized into three main paradigms, \ie, fine-grained recognition (1) with localization-classification subnetworks; (2) with end-to-end feature encoding and (3) with external information. Among them, the first and second paradigms restrict themselves by only utilizing the supervisions associated with fine-grained images such as image labels, bounding boxes, part annotations, etc. In addition, automatic recognition systems cannot yet achieve excellent performance due to the fine-grained challenges. Thus, researchers gradually attempt to involve external but cheap information (\eg, web data, text descriptions) into fine-grained recognition for further improving accuracy, which corresponds to the third paradigm of fine-grained recognition. Popularly used evaluation metric in fine-grained recognition is the averaged classification accuracy across all the subordinate categories of the datasets.

%\begin{figure}[t!]
%\centering
%\vspace{0.12em}
%	{\includegraphics[width=0.9\columnwidth]{cub_curve}}
%\caption{Recent evolution of fine-grained recognition performance. We take the classification accuracy on CUB200-2011 for illustration.}
%\label{fig:cub_curve}
%\end{figure}

\subsection{By localization-classification subnetworks}

To mitigate the challenge of intra-class variations, researchers in the fine-grained community pay attentions on capturing discriminative semantic parts of fine-grained objects and then constructing a mid-level representation corresponding to these parts for the final classification. Specifically, a localization subnetwork is designed for locating these key parts. While later, a classification subnetwork follows and is employed for recognition. The framework of such two collaborative subnetworks forms the first paradigm, \ie, fine-grained recognition with \emph{localization-classification subnetworks}.

Thanks to the localization information, \eg, part-level bounding boxes or segmentation masks, it can obtain more discriminative mid-level (part-level) representations w.r.t. these fine-grained parts. Also, it further enhances the learning capability of the classification subnetwork, which could significantly boost the final recognition accuracy.

Earlier works belonging to this paradigm depend on additional dense part annotations (\emph{aka} key points localization) to locate semantic key parts (\eg, head, torso) of objects. Some of them learn part-based detectors~\cite{Ning14ECCV,Di15CVPR}, and some of them leverage segmentation methods for localizing parts~\cite{maskcnnPR}. Then, these methods concatenate multiple part-level features as a whole image representation, and feed it into the following classification subnetwork for final recognition. Thus, these approaches are also termed as \emph{part-based} recognition methods.

However, obtaining such dense part annotations is labor-intensive, which limits both scalability and practicality of real-world fine-grained applications. Recently, it emerges a trend that more techniques under this paradigm only require image labels~\cite{STN,RACNN,MACNN,MAMCeccv} for accurate part localization. The common motivation of them is to first find the corresponding parts and then compare their appearance. Concretely, it is desirable to capture semantic parts (\eg, head and torso) to be shared across fine-grained categories, and meanwhile, it is also eager for discovering the subtle differences between these part representations. Advanced techniques, like attention mechanisms~\cite{navigateECCV18} and multi-stage strategies~\cite{XHeAAAI17FG} complicate the joint training of the integrated localization-classification subnetworks.

\subsection{By end-to-end feature encoding}

Different from the first paradigm, the second paradigm, \ie, \emph{end-to-end feature encoding}, leans to directly learn a more discriminative feature representation by developing powerful deep models for fine-grained recognition. The most representative method among them is Bilinear CNNs~\cite{TsungYu15ICCV}, which represents an image as a pooled outer product of features derived from two deep CNNs, and thus encodes higher order statistics of convolutional activations to enhance the mid-level learning capability. Thanks to its high model capacity, Bilinear CNNs achieve remarkable fine-grained recognition performance. However, the extremely high dimensionality of bilinear features still makes it impractical for realistic applications, especially for the large-scale ones.

Aiming at this problem, more recent attempts, \eg,~\cite{compactBCNN,lowrankBCNN,kernelpoolCVPR}, try to aggregate low-dimensional embeddings by applying tensor sketching~\cite{tensorKDD13,countsketch}, which can approximate the bilinear features and maintain comparable or higher recognition accuracy. Other works, \eg, \cite{maximumEntro}, focus on designing a specific loss function tailored for fine-grained and is able to drive the whole deep model for learning discriminative fine-grained representations.

\subsection{With external information}

As aforementioned, beyond the conventional recognition paradigms, another paradigm is to leverage external information, \eg, web data, multi-modality data or human-computer interactions, to further assist fine-grained recognition.

\subsubsection{With web data}

To identify the minor distinction among various fine-grained categories, sufficient well-labeled training images are in high demand. However, accurate human annotations for fine-grained categories are not easy to acquire, due to the difficulty of annotations (always requiring domain experts) and the myriads of fine-grained categories (\ie, more than thousands of subordinate categories in a meta-category).

Therefore, a part of fine-grained recognition methods seek to utilize the free but noisy web data to boost recognition performance. The majority of existing works in this line can be roughly grouped into two directions. One of them is to crawl noisy labeled web data for the test categories as training data, which is regarded as webly supervised learning~\cite{bohancvpr17,xiaxiaoaaai19}. Main efforts of these approaches concentrate on: (1) overcoming the dataset gap between easily acquired web images and the well-labeled data from standard datasets; and (2) reducing the negative effects caused by the noisy data. For dealing with the aforementioned problems, deep learning techniques of adversarial learning~\cite{gan14nips} and attention mechanisms~\cite{bohancvpr17} are frequently utilized. The other direction of using web data is to transfer the knowledge from an auxiliary categories with well-labeled training data to the test categories, which usually employs zero-shot learning~\cite{niulicvpr18} or meta learning~\cite{eccvmetalearning} to achieve that goal.

\subsubsection{With multi-modality data}

Multi-modal analysis has attracted a lot of attentions with the rapid growth of multi-media data (\eg, image, text, knowledge base, etc). In fine-grained recognition, it takes multi-modality data to establish joint-representations/embeddings for incorporating multi-modality information. It is able to boost fine-grained recognition accuracy. In particular, frequently utilized multi-modality data includes text descriptions (\eg, sentences and phrases of natural languages) and graph-structured knowledge base. Compared with strong supervisions of fine-grained images, \eg, part annotations, text descriptions are weak supervisions. Besides, text descriptions can be relatively accurately returned by ordinary humans, rather than the experts in a specific domain. In addition, high-level knowledge graph is an existing resource and contains rich professional knowledge, such as \emph{DBpedia}~\cite{DBpedia15}. In practice, both text descriptions and knowledge base are effective as extra guidance for better fine-grained image representation learning.

\begin{figure}[t!]
\centering
{\includegraphics[width=\columnwidth]{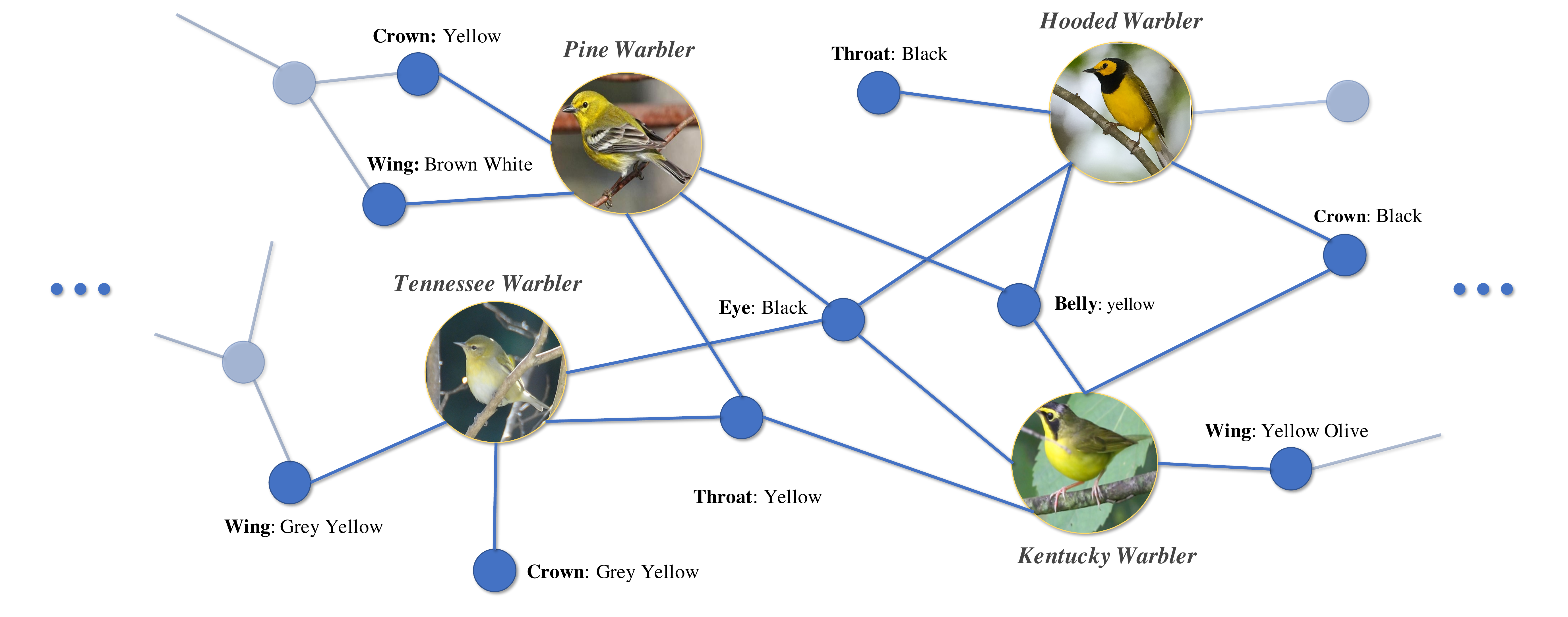}}
%\vspace{-1.6em}
\caption{An example knowledge graph for modeling the category-attribute correlations on \emph{CUB200-2011}.}
\label{fig:KB}
\end{figure}

Specifically, \cite{fgtextcvpr2016} collects text descriptions, and introduces a structured joint embedding for zero-shot fine-grained image recognition by combining texts and images. Later, \cite{yuxinpengcvpr2017} combines the vision and language streams in a joint training end-to-end fashion to preserve the intra-modality and inter-modality information for generating complementary fine-grained representations. For fine-grained recognition with knowledge base, some works, \eg,~\cite{KERLijcai18,TCNNijcai18}, introduce the knowledge base information (always associating with attribute labels, cf. Fig.~\ref{fig:KB}) to implicitly enriching the embedding space (also reasoning about the discriminative attributes for fine-grained objects).

\subsubsection{With humans in the loop}

Fine-grained recognition with humans in the loop is usually an iterative system composed of a machine and a human user, which combines both human and machine efforts and intelligence. Also, it requires the system to work in a human labor-economy way as possible. Generally, for these kinds of recognition methods, the system in each round is seeking to understand how humans perform recognition, \eg, by asking untrained humans to label the image class and pick up hard examples~\cite{yincvpr16}, or by identifying key part localization and selecting discriminative features~\cite{wisdomtpami16} for fine-grained recognition.

\section{Fine-grained image retrieval}\label{sec:fgretrieval}

Beyond image recognition, fine-grained retrieval is another crucial aspect of FGIA and emerges as a hot topic. Its evaluation metric is the common mean average precision (mAP).

\begin{figure}[t!]
\centering
{\includegraphics[width=0.95\columnwidth]{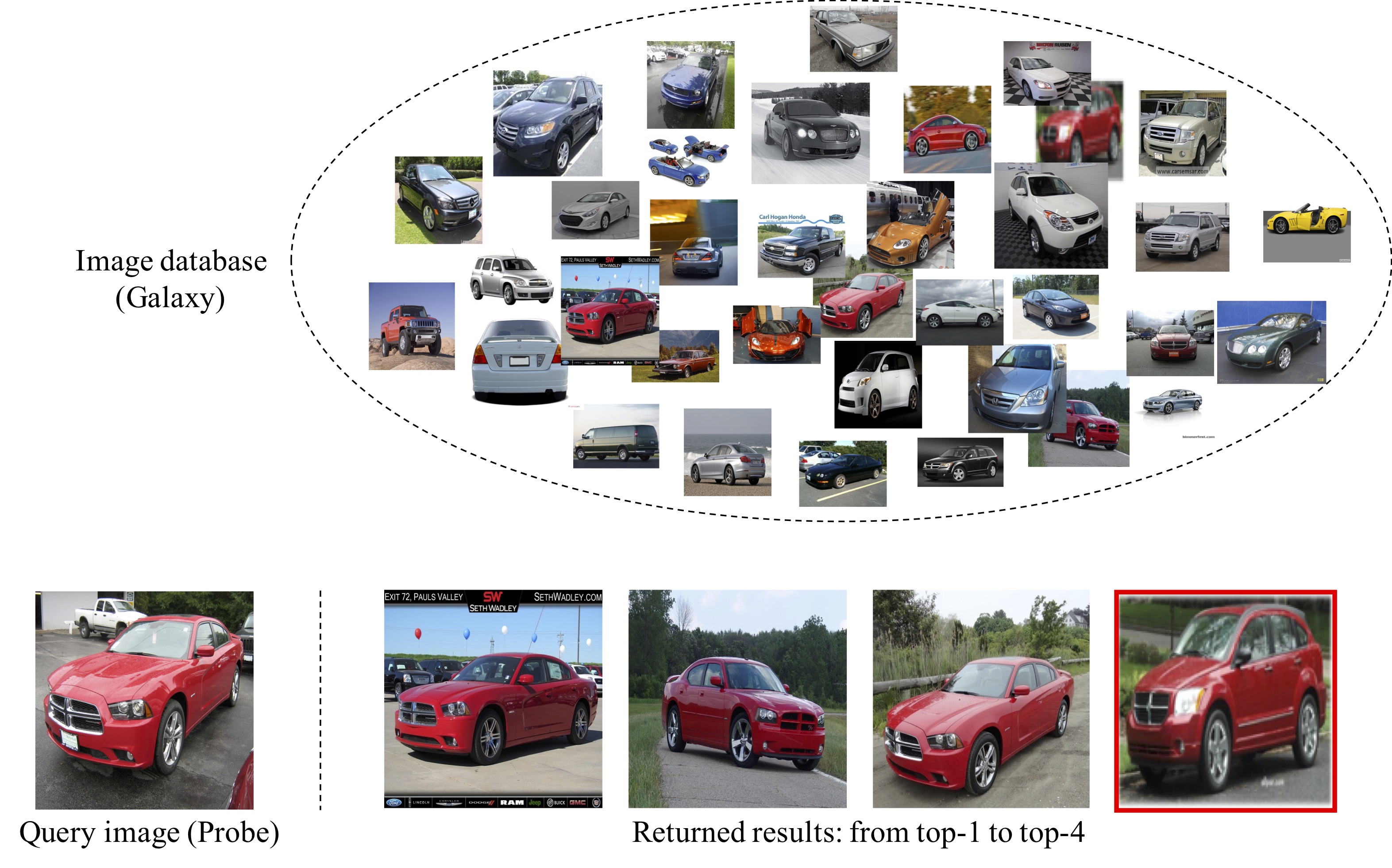}}
\vspace{-0.5em}
\caption{An illustration of fine-grained \emph{retrieval}. Given a query image (\emph{aka} probe) of ``\texttt{Dodge Charger Sedan 2012}'', fine-grained retrieval is required to return images of the same car model from a car database (\emph{aka} galaxy). In this figure, the top-4 returned image marked in a red rectangle presents a wrong result, since its model is ``\texttt{Dodge Caliber Wagon 2012}''.}
\label{fig:retrievaldemo}
\end{figure}

In fine-grained image retrieval, given database images of the same sub-category (\eg, birds or cars) and a query, it should return images which are in the same variety as the query, without resorting to any other supervision signals, cf. Fig.~\ref{fig:retrievaldemo}. Compared with generic image retrieval which focuses on retrieving near-duplicate images based on similarities in their contents (\eg, textures, colors and shapes), while fine-grained retrieval focuses on retrieving the images of the same types (\eg, the same subordinate species for the animals and the same model for the cars). Meanwhile, objects in fine-grained images have only subtle differences, and vary in poses, scales and rotations.

In the literature, \cite{Wei16scda} is the first attempt to fine-grained image retrieval using deep learning. It employs pre-trained CNN models to select the meaningful deep descriptors by localizing the main object in fine-grained images \emph{unsupervisedly}, and further reveals that selecting only useful deep descriptors with removing background or noise could significantly benefit retrieval tasks. Recently, to break through the limitation of unsupervised fine-grained retrieval by pre-trained models, some trials~\cite{xiawuijcai18,xiawuaaai19} tend to discovery novel loss functions under the \emph{supervised} metric learning paradigm. Meanwhile, they still design additional specific sub-modules tailored for fine-grained objects, \eg, the weakly-supervised localization module proposed in~\cite{xiawuijcai18}, which is under the inspiration of~\cite{Wei16scda}.

\section{Fine-grained image generation}\label{sec:fggeneration}

Apart from the supervised learning tasks, image generation is a representative topic of unsupervised learning. It deploys deep generative models, \eg, GAN~\cite{gan14nips}, to learn to synthesize realistic images which looks visually authentic. With the quality of generated images becoming higher, more challenging goals are expected, \ie, fine-grained image generation. As the term suggests, fine-grained generation will synthesize images in fine-grained categories such as faces of a specific person or objects in a subordinate category.

The first work in this line was CVAE-GAN proposed in~\cite{CVAEiccv17}, which combines a variational auto-encoder with a generative adversarial network under a conditional generative process to tackle this problem. Specifically, CVAE-GAN models an image as a composition of label and latent attributes in a probabilistic model. Then, by varying the fine-grained category fed into the resulting generative model, it can generate images in a specific category. More recently, generating images from text descriptions~\cite{AttnGANcvpr18} behaves popular in the light of its diverse and practical applications, \eg, art generation and computer-aided design. By performing an attention equipped generative network, the model can synthesize fine-grained details of subtle regions by focusing on the relevant words of text descriptions.

\section{Domain specific applications related to fine-grained image analysis}\label{sec:otherapp}

In the real world, deep learning based fine-grained image analysis techniques are also adopted to diverse domain specific applications and shows great performance, such as clothes/shoes retrieval~\cite{sketchretrievaliccv17} in recommendation systems, fashion image recognition~\cite{deepfashion16} in e-commerce platforms, product recognition~\cite{rpc} in intelligent retail, etc. These applications are highly related to both fine-grained retrieval and recognition of FGIA.

Additionally, if we move down the spectrum of granularity, in the extreme, face identification can be viewed as an instance of fine-grained recognition, where the granularity is under the \textit{identity granularity} level. Moreover, person/vehicle re-identification is another fine-grained related task, which aims at determining whether two images are taken from the same specific person/vehicle. Apparently, re-identification tasks are also under identity granularity.

In practice, these works solve the corresponding domain specific tasks by following the motivations of FGIA, which includes capturing the discriminative parts of objects (faces, persons and vehicles)~\cite{personreideccv18fg}, discovering coarse-to-fine structural information~\cite{rnnha}, developing attribute-based models~\cite{deepfashion16}, and so on.

\section{Concluding remarks and future directions}\label{sec:conclusion}

Fine-grained image analysis (FGIA) based on deep learning have made great progress in recent years. In this paper, we give an extensive survey on recent advances in FGIA with deep learning. We mainly introduced the FGIA problem and its challenges, discussed the significant improvements of fine-grained image recognition/retrieval/generation, and also presented some domain specific applications related to FGIA. Despite the great success, there are still many unsolved problems. Thus, in this section, we will point out these problems explicitly and introduce some research trends for the future evolution. We hope that this survey not only provides a better understanding of FGIA but also facilitates future research activities and application developments in this field.

\paragraph{Automatic fine-grained models} Nowadays, automated machine learning (AutoML)~\cite{automlnips} and neural architecture search (NAS)~\cite{nassurvey} are attracting fervent attentions in the artificial intelligence community, especially in computer vision. AutoML targets automating the end-to-end process of applying machine learning to real-world tasks. While, NAS, the process of automating neural network architecture designing, is thus a logical next step in AutoML. Recent methods of AutoML and NAS could be comparable or even outperform hand-designed architectures in various computer vision applications. Thus, it is also promising that automatic fine-grained models developed by AutoML or NAS techniques could find a better and more tailor-made deep models, and meanwhile it can advance the studies of AutoML and NAS in turn.

\paragraph{Fine-grained few-shot learning} Humans are capable of learning a new fine-grained concept with very little supervision, \eg, few exemplary images for a species of bird, yet our best deep learning fine-grained systems need hundreds or thousands of labeled examples. Even worse, the supervision of fine-grained images are both time-consuming and expensive, since fine-grained objects should be always accurately labeled by domain experts. Thus, it is desirable to develop fine-grained few-shot learning (FGFS)~\cite{pcmFSFG}. The task of FGFS requires the learning systems to build classifiers for novel fine-grained categories from few examples (only one or less than five) in an meta-learning fashion. Robust FGFS methods could extremely strengthen the usability and scalability of fine-grained recognition. 

\paragraph{Fine-grained hashing} As there exist growing attentions on FGIA, more large-scale and well-constructed fine-grained datasets have been released, \eg,~\cite{Birdsnap14,inat2017,rpc}. In real applications like fine-grained image retrieval, it is natural to raise a problem that the cost of finding the exact nearest neighbor is prohibitively high in the case that the reference database is very large. Hashing~\cite{surveyhashtpami,wujunhashingijcai16}, acting as one of the most popular and effective techniques of approximate nearest neighbor search, has the potential to deal with large-scale fine-grained data. Therefore, fine-grained hashing is a promising direction worth further explorations.

\paragraph{Fine-grained analysis within more realistic settings} In the past decade, fine-grained image analysis related techniques have been developed and achieve good performance in its traditional settings, \eg, the empirical protocols of~\cite{WahCUB200_2011,Khosla11stanforddogs,cars}. However, these settings can not satisfy the daily requirements of various real-world applications nowadays, \eg, recognizing retail products in storage racks by models trained with images collected in controlled environments~\cite{rpc} and recognizing/detecting natural species in the wild~\cite{inat2017}. In consequence, novel fine-grained image analysis topics, to name a few---fine-grained analysis with domain adaptation, fine-grained analysis with knowledge transfer, fine-grained analysis with long-tailed distribution, and fine-grained analysis running on resource constrained embedded devices---deserve a lot of research efforts towards the more advanced and practical FGIA.

\clearpage
{%\footnotesize
\scriptsize
\bibliographystyle{named}
\bibliography{ijcai19_finegrained_survey}
}

\end{document}